\documentclass[preprint, 3p, authoryear]{elsarticle}
\usepackage[utf8]{inputenc}
\usepackage{amssymb,amsfonts,amsmath}
\usepackage{graphicx}
\usepackage{multirow}
\usepackage{color}
\usepackage[kernelfbox]{xcolor}

\usepackage{float}
\usepackage{wrapfig}
\restylefloat{figure}
\usepackage{booktabs}
\usepackage[english]{babel}

\usepackage{caption}
\usepackage{subcaption}
\usepackage{csquotes}

\usepackage{setspace}
\usepackage{natbib}
\usepackage{array}
\usepackage{calc}

\DeclareMathOperator*{\argmax}{arg\,max~}

\usepackage[colorlinks]{hyperref}
\usepackage{chngcntr}

\usepackage[nameinlink,capitalize]{cleveref}
\crefformat{appendix}{#2#1#3}

\newcommand{\w}{\mathbf{w}}
\newcommand{\x}{\mathbf{x}}

\newcommand{\z}{\mathbf{z}}

\newcommand{\me}{\mathbf{c}}
\newcommand{\out}{\mathbf{o}}

\newcommand{\h}{\mathbf{h}}

\newcommand{\A}{\mathbf{A}}
\newcommand{\B}{\mathbf{B}}
\newcommand{\E}{\mathbf{E}}
\newcommand{\U}{\mathbf{U}}

\newcommand{\W}{\mathbf{W}}

\usepackage{lineno}
\modulolinenumbers[5]

\journal{Journal of Visual Communication and Image Representation}
\begin{document}

\begin{frontmatter}

\title{Egocentric Video Description based on Temporally-Linked Sequences}

\author[ub_address,cvc_address]{Marc Bola\~nos}
\ead{marc.bolanos@ub.edu}
\author[upv_address]{\'Alvaro Peris}
\ead{lvapeab@prhlt.upv.es}
\author[upv_address]{Francisco Casacuberta}
\ead{fcn@prhlt.upv.es}
\author[ub_address]{Sergi Soler}
\ead{ssolerso8@alumnes.ub.ed}
\author[ub_address,cvc_address]{Petia Radeva}
\ead{petia.ivanova@ub.edu}

\address[ub_address]{Universitat de Barcelona, Barcelona, Spain}
\address[cvc_address]{Computer Vision Center, Bellaterra, Spain}
\address[upv_address]{PRHLT Research Center, Universitat Polit\`ecnica de Val\`encia, Val\`encia, Spain}

\begin{abstract}
Egocentric vision consists in acquiring images along the day from a first person point-of-view using wearable cameras. The automatic analysis of this information allows to discover daily patterns for improving the quality of life of the user. A natural topic that arises in egocentric vision is storytelling, that is, how to understand and tell the story relying behind the pictures. 

In this paper, we tackle storytelling as an egocentric sequences description problem. We propose a novel methodology that exploits information from temporally neighboring events, matching precisely the nature of egocentric sequences. Furthermore, we present a new method for multimodal data fusion consisting on a multi-input attention recurrent network. We also release the \textit{EDUB-SegDesc} dataset. This is the first dataset for egocentric image sequences description, consisting of 1,339 events with 3,991 descriptions, from 55 days acquired by 11 people.
Finally, we prove that our proposal outperforms classical attentional encoder-decoder methods for video description.

\end{abstract}


\end{frontmatter}


\section{Introduction} \label{sec:introducion}

 \begin{wrapfigure}{r}{0.55\textwidth}
    \vspace{-1em}
	\centering
	\includegraphics[width=0.55\textwidth]{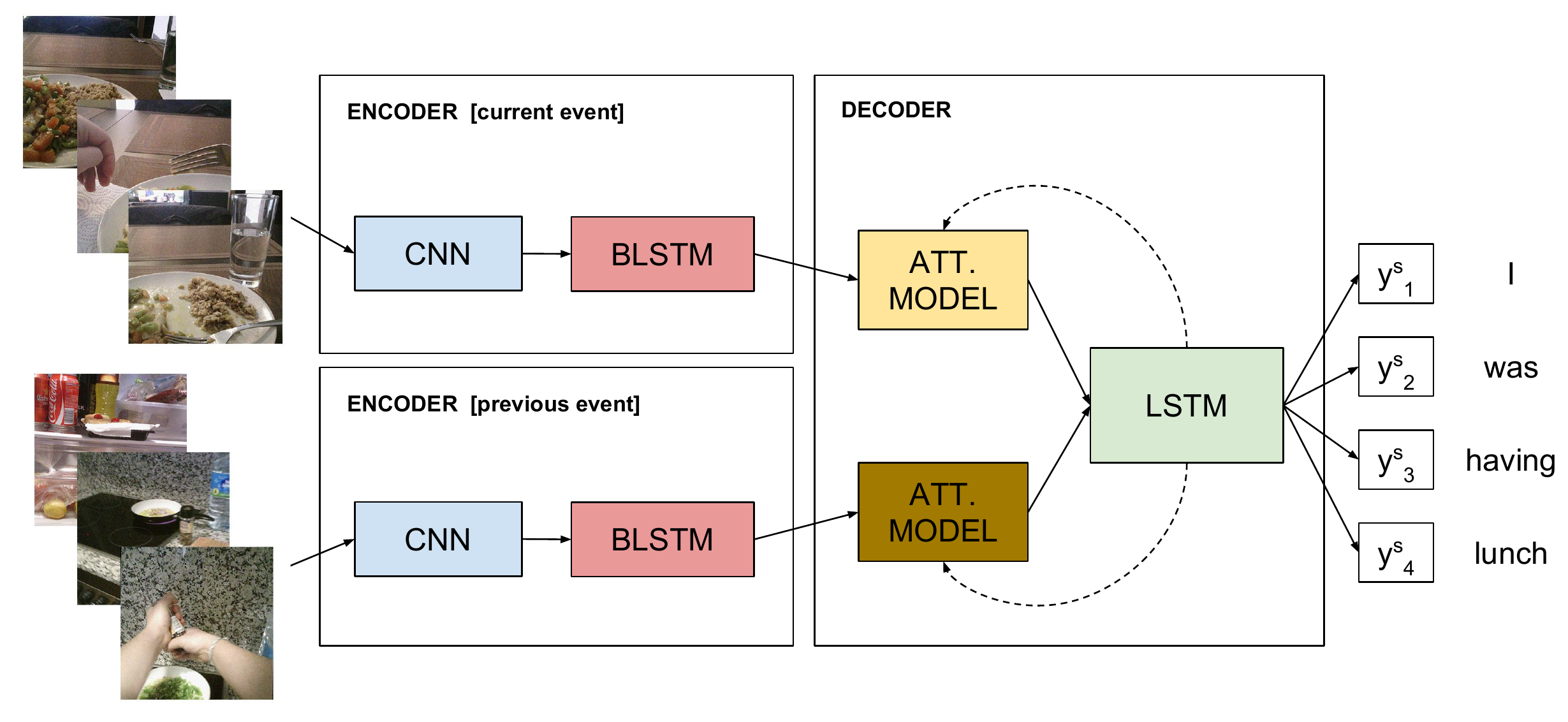}
	\caption{\label{fig:TMA_outline} General outline of the proposed temporally-linked multi-input attention model (TMA).}
\end{wrapfigure}

Egocentric vision \citep{doherty2013wearable,betancourt2015evolution,bolanos2016toward} is a recent topic in the computer vision field, with the goal of analyzing the visual information provided by wearable cameras, which have the capability to acquire images or videos from a first person point-of-view. The analysis of these images can provide useful information about the behavior of the user for several complementary topics like social interactions, scene understanding, time-space-based localization, action recognition, nutritional habits, among others. Thus, enabling to understand the whole story and user's behavior behind the pictures (i.e. automatic storytelling) followed by inferring his/her actions and habits, could lead to a better quality of life for the user. Considering the sheer amount of data wearable cameras provide, there is a need to create automatic algorithms to analyze and summarize them. In this paper, we focus on the specific topic of creating automatic diaries of the life of the user by means of textual descriptions. One of the possible health-related applications for the automatic diary construction could be the treatment of patients with dementia. As proven by \citet{spector2003efficacy,sellen2007life}, the daily review of egocentric pictures taken by this kind of patients can help them recover partially their cognitive capabilities. The incorporation of additional automatically extracted textual information and comparing it to the user's one could give novel tools to complete cognitive frameworks for memory enhancement.

The egocentric captioning problem can be seen as an instantiation of a video description task \citep{venugopalan2015sequence}: the system receives as input a sequence of images and must produce a sentence as a sequence of words describing it. The problem is challenging due to two main reasons: on the one hand, most wearable cameras have a small field of view the other hand, lifelogging cameras have low temporal resolution (2-3 fpm) that is events are not videos, but collection of temporally ordered images. Hence, the problem is much more difficult than conventional videos and we need powerful techniques and algorithms from computer vision and natural language understanding to address the problem of video description. The recent development of deep learning techniques has allowed a breakthrough in the computer vision and natural language processing fields. The ability of training complex models has pushed forward many research areas and nowadays, Convolutional Neural Networks (CNNs) constitute one of the most powerful tools in the computer vision field. As the problem at hand involves sequence learning or prediction, a natural choice are Recurrent Neural Networks (RNNs), which are powerful sequence modelers. Specifically, the use of gated units, such as Long Short-Term Memory (LSTM)~\citep{Hochreiter97} or gated recurrent units~\citep{Cho14} allows to properly model sequences with long and complex relationships.

The combination of RNNs together with CNNs is therefore widely employed for tackling multi-modal learning tasks, involving the combination of vision and language. Some examples of multi-modal learning problems are image or video description, dense captioning, visual question answering, multi-modal interaction to mention a few.

In this work, we develop a fully-neural end-to-end system for egocentric captioning that we call Temporally-linked Multi-input Attention model (TMA). We hypothesize that, within the egocentric sequences captioning problem, given a day, some of the events that compose it can follow a temporally logical relation. That is, previous actions occurred during a day can influence the following ones. Therefore, we need a model able to capture and learn this relation.
Our proposal is able to embed previous information coming from either image or language. In \cref{fig:TMA_outline}, we show a simplified outline of the proposed method. 
In this example, we illustrate how previous event's frames can be employed in order to help to the process of video description of the analyzed event. Note that in the example, the user is cooking and next he proceeds to take the food. This temporal relationship between actions aids the model when predicting the current description.

Furthermore, we publish the first egocentric dataset for event captioning, composed of 55 complete days containing nearly 55,000 images in total, acquired by 9 different people. 

The main contributions of this work are the following:

\begin{itemize}
\item We present a novel captioning model, which incorporates the information from previous events into the current decoding state.
\item We present a new LSTM model capable of combining information from multiple inputs and modalities, as well as applying a separate attention mechanism to each of them.
\item We conduct experiments on the new dataset and compare our model with other classical captioning architectures. Results show that using information from previous events provides better generalization. 
\item We present the first dataset for egocentric sequences captioning, named EDUB-SegDesc, based on describing the events that take place along a day.
\item We make public both dataset and model, in order to make results reproducible and foster the research in this topic.
\end{itemize}

The rest of the paper is structured as follows:  the related work is reviewed in~\cref{sec:related-work}. Next, we describe our model and its main components in~\cref{sec:methodology}. We present the egocentric captioning dataset in~\cref{sec:dataset}. We set up our experimental framework and show the obtained results in~\cref{sec:results}. Such results are analyzed in~\cref{sec:discussion}. Finally, conclusions and future research lines are drawn in~\cref{sec:conclusions}.
\section{Related work} 
\label{sec:related-work}

In recent years, deep learning techniques have provided tremendous advances in the computer vision field. More precisely, CNNs~\citep{Krizhevsky12} have proved to excel on the task of learning rich image representations and consequently have served as feature extractors, providing record-breaking results in most tasks. 
On the other hand, RNNs have proved to be powerful sequence modelers. They have been recently used in many sequences learning tasks, including machine translation~\citep{Sutskever14,Bahdanau15}, image~\citep{Xu15} and video~\citep{Yao15,Peris16} captioning, or visual question answering~\citep{fukui2016multimodal,bolanos2016vibiknet}. Most of these problems involve tackling multi-modal data (i.e. text and images) \citep{Toselli11}, which means that, in all of them, both CNNs and RNNs are commonly employed. 

Such systems are built under the encoder-decoder framework: an encoder processes the input and computes a meaningful representation of it. Next, the decoder takes this representation and produces the desired output, typically a sentence in natural language. Given their power as feature extractors, CNNs are usually employed as encoders~\citep{Xu15,Yao15,fukui2016multimodal}. Nevertheless, if the input signal has a sequential component (as in machine translation or video captioning), RNNs, solely or together with CNNs, can also act as encoders~\citep{Sutskever14,Peris16}.

The vanilla encoder-decoder architecture was enhanced with attention mechanisms \citep{Bahdanau15,Xu15,Yao15}, allowing the decoder to selectively focus on parts of the input during the text generation process. Therefore, the system is able to properly deal with long and complex inputs.

Regarding the video description problem, which consists in generating a natural language description given an input video or sequence of images, several proposals have been published in recent years. The general framework used also consists of applying one or several CNNs to the input images, followed by a generative LSTM for sentence generation. Several improvements and variations have been proposed to the main model, some examples being the use of additional CNN representations like optical flow \citep{venugopalan2015sequence,Yao15}, Bidirectional LSTMs (BLSTM) in the encoder \citep{Peris16}, attention mechanisms in the decoder \citep{Yao15}, hierarchical information \citep{pan2016hierarchical}, external linguistic knowledge \citep{venugopalan2016improving} or multi-modal attention mechanisms \citep{Hori17}, among others.

All these works relating video captioning assume a sample-wise independence. This is that a sample is meant to be unrelated to the next one. This may represent a limitation in tasks which aim to model continuous events, arbitrarily split. The task addressed in this work belongs to this class of problems: we aim to describe the whole day of a user. For tackling it, we divide a day into events and describe each one of these events. Due to this division, our samples are conditioned between them. For example, if the user is at the office, it is likely that in the following event he/she uses a computer. Therefore, the classical sample-wise independence assumption becomes excessively severe, as critical information may be potentially lost. We propose a novel model that takes into account information from past events, being suitable for this kind of tasks.

Only one work has recently proposed to take into account mid-term temporal information in conventional videos \citep{krishna2017dense}. Their model extracts C3D features on variable-sized conventional videos and computes several temporal segmentation proposals in short actions. Later, it generates textual descriptions for each action incorporating contextual information from past and future actions belonging to the same video. In comparison to our problem of egocentric lifelogging data, apart from the different perspective of the sequences, their videos are much shorter. Moreover, instead of consisting of several long-term events, theirs contain mid-term short actions belonging to the same event. Even though there are important differences to both problems, the authors prove that incorporating past and future information helps predicting the output of current events or actions.

In the field of egocentric vision \citep{doherty2013wearable,betancourt2015evolution,bolanos2016toward}, some authors have worked on related problems like activity recognition \citep{iwashita2014first,cartas2017batch} or event classification \citep{castro2015predicting}. The main handicap of these methods is that they provide simple labels from a user-oriented perspective, which provide more limited and predefined information compared to generating free textual descriptions. Regarding any of these methods, the first step needed in order to provide a coherent description of the actions and events happening in egocentric images is the application of an automatic segmentation \citep{dimiccoli2016sr,poleg2016compact,lu2013story} of the complete day of the user. After the application of the day segmentation and having as output the set of events of the day, we can acquire the self-contained units of information needed in order to apply either event captioning, activity recognition or event classification models.

Considering the specific problem at hand, which is the generation of textual descriptions of egocentric events, few works have been proposed in the state-of-the-art. Only one of them tackled it as a video description problem and from an end-to-end perspective. \citet{fan2016deepdiary} explored the problem of creating image diaries in order to apply image retrieval. As a step of its process, the authors proposed an image captioning method that processes one image at a time by applying a CNN for features extraction and a RNN for sentence generation. Later, they proposed grouping the images and applying a sentences fusion technique for providing the final captions for each event along the day. \citet{goeldeepseek} also focused on the task of image retrieval for both conventional and egocentric videos. It applied a simple method of video description composed of a CNN for feature extraction, a Bidirectional LSTM for image sequence combination and a LSTM in the decoder for applying the final sentence generation for 5-seconds-long clips of video. The purpose of this method in the work was to provide semantic information of the available data for the posterior retrieval.


In this work, we focus on event textual description. We assume that the events are previously extracted, manually or automatically~\citep{dimiccoli2016sr}. To this aim, we propose an end-to-end model specifically trained for egocentric day sequences captioning.  As argued before, it is important to take into account relevant information from previous samples, for generating descriptions of the current event. Thus, our model treats the image sequences from different events as temporally-linked units. It jointly models and exploits intra-eventual information (flowing through the frames of a single event) together with inter-eventual information (linking temporal sequences of neighboring events). 
\section{Methodology} 
\label{sec:methodology}
The problem, we face in this work,  involves a multi-modal learning task. On the one hand, we have a huge amount of information coming from the sequence of images captured by the wearable camera. On the other hand, each sequence of images has the correspondent caption associated, which describes the event occurring in the life of the user from an egocentric point of view.

Although the sequences of images may be segmented in independent semantic units to some extent (i.e. events), given the nature of the egocentric images, it is obvious that there exists a dependency between temporally neighboring events (i.e. most given events experienced in the day of a person have a relationship to the previous one he/she lived). We aim to exploit this information by incorporating into our model, at a given temporal point $s$, the information extracted from the previous event ${s-1}$. This information can be either textual (previous caption), visual (previous sequence of images) or both. Therefore, we develop a model that takes into account the information from both sources: the current sequence of frames together with the information coming from previous events.

\subsection{Egocentric captioning}

\begin{figure}[!ht]
	\centering
	\includegraphics[width=0.94\textwidth]{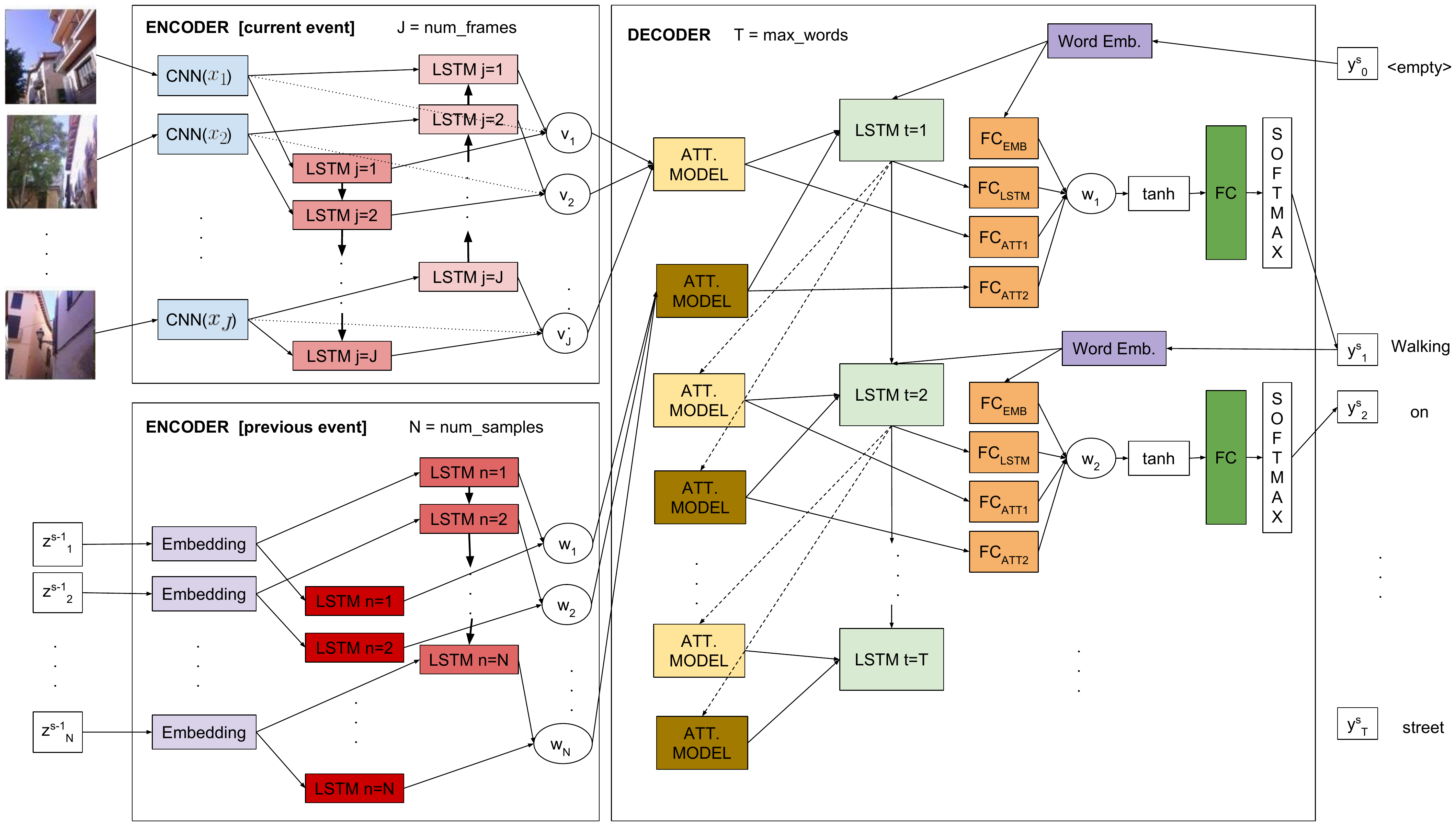}
	\caption{\label{fig:temporally_linked} Architecture of the proposed TMA model. Unlike the traditional approaches, our architecture consists of at least two encoder stages; one for the current sequence and another one (or more) for the previous sequence; and a decoder stage that combines the information of all the previous stages using a multi-input attention LSTM.}
	\vspace*{0.2cm}
\end{figure}  

Similarly to the video description problem, in the egocentric task, we have as input a sequence of frames and we want to output a sentence that describes the input. This problem has already been tackled in the literature on conventional videos with multiple variations~\citep{pan2016hierarchical,venugopalan2015sequence,venugopalan2016improving,Yao15,Peris16}. In the latter work, 
the frames were encoded by a CNN and a BLSTM network. The representation obtained was fed to an LSTM decoder equipped with an attention mechanism, which generated the corresponding caption. We use this architecture as starting point for developing our system.

The main difference that characterizes the egocentric captioning problem is that, unlike in the classical video description approach, we do have temporally-linked events, which share a relationship (e.g. if in an event the user enters into an office and sits on his/her table, it is highly likely that in the following event he/she will be using a computer). 

Thus, we propose a system able to take advantage from both the current sequence of egocentric images and the action happened in the previous event. In \cref{fig:temporally_linked} we detail the architecture of our model. The input to the system is a sequence of frames $X^s = x_1, \dotsc, x_J$ and the sequence of information providing from the previous event, $Z^{s-1} = z^{s-1}_1, \dotsc, z^{s-1}_N$. This latter sequence can either be the previous textual description, $Y^{s-1}$ or the previous sequence of frames, $X^{s-1}$, or both. The current sequence of images is processed in the encoder of the current event, which is composed of a CNN (blue) and of a BLSTM (light red), producing the sequence of visual features $\mathbf{v}_1, \dotsc, \mathbf{v}_J$. Meanwhile, the data $Z^{s-1}$ is processed in the encoder of the previous event, which is composed of an embedding mechanism (light purple) and of another BLSTM (dark red), which computes the final sequence of representations $\w_1, \dotsc, \w_N$. The embedding mechanism used in this encoder can either be a CNN (blue) when dealing with images or a word embedding matrix (dark purple) when dealing with a textual representation. Note that this word embedding matrix is the same as the one in the decoder. These sequences of feature vectors are given as input to the decoder LSTM by means of two independent attention mechanisms (yellow and brown). In addition, at each time-step $t$, the decoder takes as input the word embedding of the previously generated word (${\E}(y^s_{t-1})$). Finally, a series of skip connections is combined in an element-wise summation after embedding the different modalities by using fully-connected layers (orange). We include a skip connection for each attention mechanism, one for the LSTM hidden state and another one for the previously generated word. The sequence of time-steps for the current caption is indexed by $t$, while the sequence of events is indexed by $s$.

We extend the classical encoder--decoder approach by adding different encoders: one for the current sequence of images and another for the information from previous events. Such encoders are built using CNNs and BLSTMs networks, which we detail in \cref{sec:encoders}. In \cref{sec:mutliLSTM}, we proceed to explain the decoder, which is also extended by adding a multi-input attention LSTM for dealing with input sequences of multiple modalities. Finally, in \cref{sec:TMA}, we summarize the whole pipeline applied in the proposed TMA model.

 \subsection{Encoder}
 \label{sec:encoders}
 
\subsubsection{Convolutional Neural Networks}
 CNNs have proved to be great modelers of image representations. Several models have been proposed in the state-of-the-art (e.g. AlexNet \citep{Krizhevsky12}, VGG \citep{simonyan2014very}, GoogLeNet \citep{Szegedy15}, ResNet \citep{he2016deep}, etc.), most of them originally for the problem of object recognition or detection \citep{russakovsky2015imagenet}. Furthermore, they have been proven to serve as excellent feature extractors for other related tasks in the computer vision field.
 
Without loss of generality, in our proposal, we make use of the well-known GoogLeNet architecture  \citep{Szegedy15} pre-trained on the ILSVRC challenge data as an extractor for training our complete TMA model. Although it varies depending on the problem and data, GoogLeNet has proven to be one of the best models for feature extraction, offering a trade-off between performance, number of parameters and computational time. We have to note that it is possible to use any alternative state-of-the-art CNN model for dealing with the images representation in our TMA model.
 
\subsubsection{Long Short-Term Memory and Bidirectional networks} 

Since the input of our system is a sequence (of images), we can effectively model it and its relationships by using a RNN. More precisely, in order to learn a representation of the input sequences and its relationships, we use a LSTM network~\citep{Hochreiter97,Gers00}. Unlike simple units, LSTM cells feature an additional state, the so-called memory state. The network controls how information flows through the unit by means of three gates, namely input, output and forget gates. Such gates modulate the amount of information that the network will incorporate from the input, from the output at the current time-step, or the amount that it will store for future time-steps. Therefore, LSTM networks are able to model long and complex sequences of information, reducing the vanishing gradient problem~\citep{Bengio94}. 

\begin{wrapfigure}{r}{0.45\textwidth}
	\centering	\includegraphics[width=0.45\textwidth]{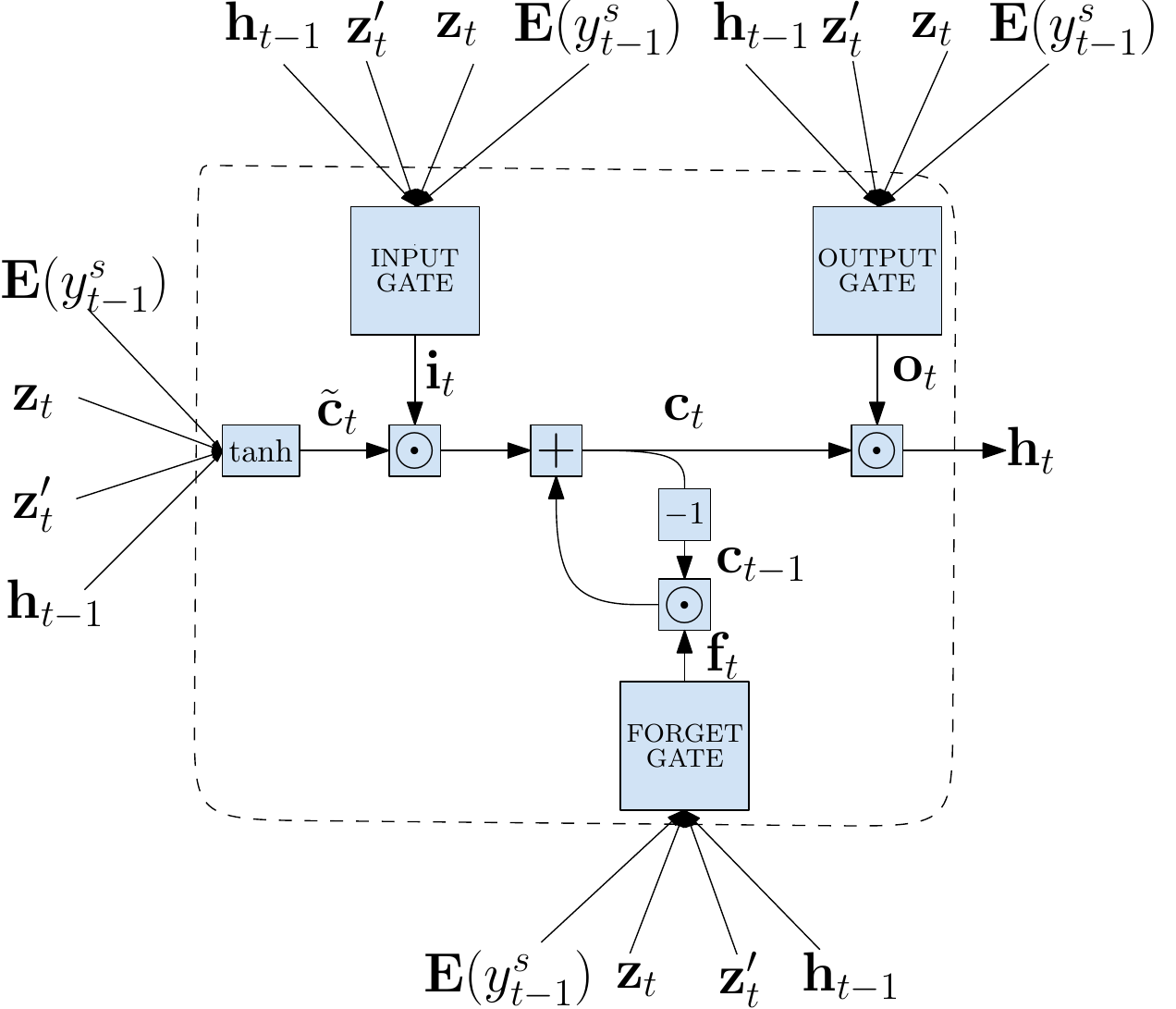}
	\caption{\label{fig:LSTM} Architecture of a multi-input LSTM cell at time-step, $t$. In this case, the inputs to the cell are the word embedding of the previous word ($\E(y^s_{t-1})$), the previous LSTM hidden state, (${\h}_{t-1}$) and the context vectors from two different attention mechanisms, ($\z_{t}, \z'_{t}$).}
\end{wrapfigure} 
 
 Hence, at a given time-step $j$, the hidden state $\mathbf{v}_j$ is the memory state $\me_j$ controlled by the output gate ${\out}_j$. The memory state depends on the previous memory state (modulated by the forget gate $\mathbf{f}_j$) and an updated memory state (modulated by the input gate $\mathbf{i}_j$). The updated state is obtained from the input $\x_j$ and the previous hidden state $\mathbf{v}_{j-1}$. All three gates are also computed from $\x_j$ and  $\mathbf{v}_{j-1}$~\citep{Gers00}.

 With LSTM units, we manage how information flows through the network, and we are able to model long-term relationships. Nevertheless, this information only flows in a time direction. If we have the full day to analyze, we can take profit from relationships from the previous to the next events, but also from information flowing from next to previous events. Bidirectional networks~\citep{Schuster97} cope with both sequence directions by maintaining two independent  recurrent layers: the forward layer processes the sequence from the past to the future and the backward layer processes the sequence reversed in time. Therefore, we can extract relationships flowing in both time directions of our input signal. Once the full sequence has been processed in both directions, forward and backward layers are combined, typically by concatenating their hidden states~\citep{Bahdanau15,Peris16}.

\subsection{Decoder: Multi-input Attention LSTM}
\label{sec:mutliLSTM}

Since we are in a multi-modal scenario dealing with sequences of text and images, we propose to use an LSTM network that accepts multiple inputs---from multiple sources---and combines them after applying an attention mechanism for each input. The combination, as well as the attention mechanisms, are thus learned together with the rest of the model. 

Attention mechanisms~\citep{Bahdanau15,Xu15} help the decoder to selectively focus on parts of the input sequence, depending on the decoding step. More formally, given a sequence of $J$ input vectors ${\bf v}_1, \dots, {\bf v}_J$ previously calculated by the encoder, at each decoding time-step $t$, the attention mechanism weights them into a single context vector $\z_t$:
\begin{equation}
\z_t = \sum_{j=1}^J \alpha_{j t} {\bf v}_j,
\end{equation}
being $\alpha_{j t}$ the weight given to the $j$-th feature vector at time, $t$. At each time-step, the weights are calculated according to the  scores obtained by a soft-alignment model, which determines how much a feature vector does influence the outcome of the current word. The aligner is implemented by a single-layer perceptron:
\begin{equation}
e_{j t} = \w^\top \tanh(\W_a \h_{t-1} + \U_a {\bf v}_j),
\end{equation}
where $\w$, $\W_a$ and $\U_a$ are the perceptron parameters and $\h_{t-1}$ is the decoder hidden state from the previous time-step. For clarity, we omit the bias term in this equation, as well as in the rest of the paper. The aligner is then followed by a softmax function in order to obtain the final normalized weights, $\alpha_{jt}$:
\begin{equation}
\alpha_{j t} = \frac{\exp{(e_{j t})}}{\sum_k^J\exp{(e_{k t})}}. 
\end{equation}

Our multi-input decoder is a natural extension of classical LSTM networks, which is able to process $N$ different inputs. For simplicity, in this section we assume that we have inputs from two different modalities. Such inputs are processed by the respective attention models. Hence, at a given time-step, $t$, we have two inputs from the attention models ($\z_t$ and $\z'_t$). Furthermore, in order to boost the decoder capabilities, we introduce the word embedding of the previously generated word as an additional input. \cref{fig:LSTM} illustrates the LSTM cell. 

As in regular LSTMs, in our multi-input network, the hidden state depends on the memory state and the output gate:
\begin{equation}
{\h}_t = {\bf o}_t \odot {\bf c}_t,
\end{equation}
where $\odot$ denotes the element-wise multiplication. ${\bf c}_t$ is the memory state, defined as:
\begin{equation}
{\bf c}_t = {\bf f}_t \odot {\bf c}_{t-1} + {\bf i}_t \odot {\tilde {\bf c}}_t,
\end{equation}
where ${\bf c}_{t-1}$ is the previous time-step memory state and ${\tilde {\bf c}}_t$ is the updated memory state. ${\bf f}_t$ and ${\bf i}_t$ are the forget and input gates, which modulate them. ${\tilde {\bf c}}_t$ is computed taking into account the attended representations of the inputs ($\z_t$ and $\z'_t$), the last word generated by the decoder ($y_{t-1}$) and the previous hidden state ($\h_{t-1}$):
\begin{equation}
{\tilde {\bf c}}_t = \tanh({\W}_c{\E}(y_{t-1}) + {\U}_c{\h}_{t-1} + {\A}_c\z_{t} + {\B}_c\z'_{t}), 
\end{equation}
where ${\E}$ is a word embedding matrix and ${\E}(y_{t-1})$ denotes the word embedding of the previously generated word. $\W_c$, $\U_c$, $\A_c$ and  $\B_c$ are the weight matrices for the word embedding of $y_{t-1}$, the previous hidden state and both context vectors from the alignment attention models, respectively.

The forget, input and output gates also depend on the context vectors, the previous word and the previous hidden state: 
\begin{equation}
{\bf f}_t = \sigma({\W}_f{\E}(y_{t-1}) + {\U}_f {\h}_{t-1} +  {\A}_f\z_{t} +  {\B}_f\z'_{t}   ),
\end{equation}
\begin{equation}
{\bf i}_t = \sigma({\W}_i{\E}(y_{t-1}) + {\U}_i {\h}_{t-1} + {\A}_i\z_{t}  +  {\B}_i\z'_{t}), 
\end{equation}
\begin{equation}
{\bf o}_t = \sigma({\W}_o{\E}(y_{t-1}) + {\U}_o {\h}_{t-1} +  {\A}_o\z_{t} +  {\B}_o\z'_{t} ).
\end{equation}

\subsection{Temporary-linked Multi-input Attention model}\label{sec:TMA}

Considering our TMA model as a whole, first we process the sequence of images captured by the camera for the current event $s$, $X^s= x^s_1, \dotsc, x^s_J$. As image encoder, we use a CNN that extracts features from $X^s$. Next, we apply a BLSTM network, in order to capture the temporal relationships existing in the sequence. Finally, for each frame $x^s_j$, we compute a feature vector $\mathbf{v}_j$ by concatenating the representation extracted by the CNN together with the forward and backward hidden states from the BLSTM.

Simultaneously, we process the previous event ($Z^{s-1}$). The information of the previous event can be either the previous sequence of egocentric images ($X^{s-1}$), the previous caption ($Y^{s-1}$), or both. In case of processing the previous sequence of images, the method is the same as the one applied on the current event: a CNN combined with a BLSTM network. 
If the information providing from the previous event is the caption $Y^{s-1} = y^{s-1}_1, \dotsc y^{s-1}_N$, words are projected to the continuous space by means of the embedding matrix, shared with the decoder. Next, an additional BLSTM processes this sequence of word embeddings. We concatenate the forward and backward BLSTM hidden states for obtaining a contextual representation for each input word. 
In case, we are using both modalities, the model is modified for having three different encoders: 1) current event frames $X^{s}$, 2) previous event frames $X^{s-1}$ and 3) previous event output caption $Y^{s-1}$, as well as three attention mechanisms in the decoder, one for each input.
For the sake of clarity and without loss of generality, let us assume from now on that we only have one modality as additional input from the previous event. 

Therefore, our encoder produces two sequences of features: $\mathbf{v}_1, \dotsc \mathbf{v}_J$, referring to the current video and $\w_1, \dotsc, \w_N$, related to the previous event. We integrate this into the decoder by means of two independent attention mechanisms. Each attention mechanism weights the elements of its respective sequence and computes a joint representation of it, $\z_{t}$ and $\z'_{t}$ respectively, taking into account the previous decoding state ${\h}_{t-1}$ (see~\cref{sec:mutliLSTM}). Such representations, together with the previously generated word, are the inputs of the multi-input LSTM (described in~\cref{sec:mutliLSTM}).

Finally, at each time-step $t$, and in order to predict the following word in the output sequence, we define a probability distribution over the task vocabulary as follows:
\begin{equation}
{\bf p}_{t} = \textsf{softmax} ({\U}_p\tanh({\W}_a{\h}_{t} + {\W}_b \z_{t} + {\W}_c \z'_{t} + {\W}_d {\E}(y^s_{t-1}))),
\end{equation}
where $\h_{t}$ is the hidden state from the decoder, $\z_{t}$ and $\z'_{t}$ are the contextual representations of the current and previous inputs computed by their respective attention models and ${\E}(y^s_{t-1})$ is the word embedding of the preceding word in the current sequence. ${\W}_a$, ${\W}_b$, ${\W}_c$, ${\W}_d$ are the skip-connections weight matrices (orange in~\cref{fig:temporally_linked}). ${\U}_p$ is the last layer matrix (dark green in \cref{fig:temporally_linked}). 

The produced distribution, ${\bf p}_t$ represents the probability of a word given the input video $X^{s}$, the previous event $Z^{s-1}$, and the words generated so far in the current sequence $y^s_{1}, \dots, y^s_{t-1}$:
\begin{equation}
{\bf p}_t = p(y^s_t|X^{s}, Z^{s-1}, y^s_{1}, \dots, y^s_{t-1}).
\end{equation}

We approximate the most likely caption by using a beam search method~\citep{Sutskever14}. 
The complete model parameters ($\theta$) are jointly estimated over a dataset $\cal S$, which consists of $S$ image sequences and caption pairs. 
The training objective is to maximize the log-likelihood of ${\cal S}$ with respect to $\theta$:
\begin{equation}
\hat{\theta} = \argmax_\theta \sum_{s=1}^S \sum_{t=1}^{T_s} \log \left( p(y^s_t|X^{s}, Z^{s-1}, y^s_{1}, \dots, y^s_{t-1}; \theta  \right),
\end{equation}
where $T_s$ is the length of the $s$-th caption. If the previous event, $Z^{s-1}$ of a given image sequence $X^{s}$ is undefined, we introduce an artificial empty event as $Z^{s-1}$.

\section{EDUB-SegDesc dataset} 
\label{sec:dataset}

\begin{figure}[!ht]
	\centering
	\includegraphics[width=0.6\textwidth]{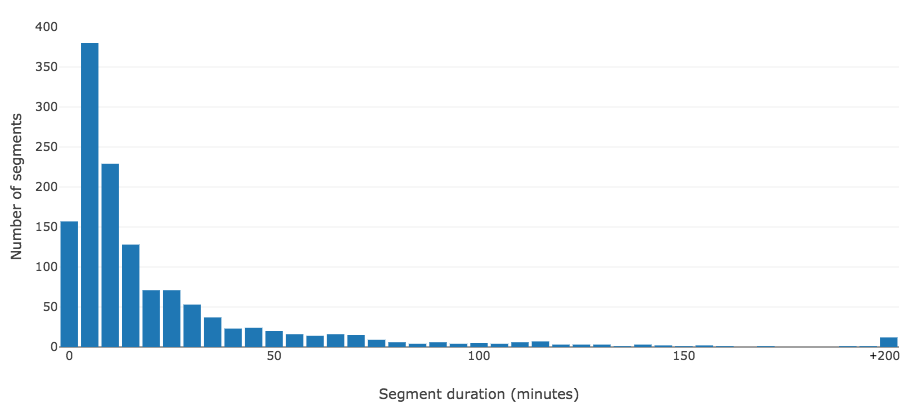}
	\caption{\label{fig:duration} Histogram depicting the duration of the segments in the EDUB-SegDesc dataset. All segments with a duration equal or greater than 200 minutes appear grouped in the last bin (+200).}
	\vspace*{0.2cm}
\end{figure}

EDUB-SegDesc\footnote{\url{http://www.ub.edu/cvub/edub-segdesc}} is a dataset that can be used either for egocentric events segmentation \citep{dimiccoli2016sr} or for egocentric sequences description. It was acquired by the wearable camera Narrative\footnote{\url{www.getnarrative.com}}, taking a picture every 30 seconds (2 fpm). It consists of 55 days acquired by 9 people. Each day was manually segmented in events or sequences following the same criteria as in \citet{dimiccoli2016sr}:
\blockquote{An event is a semantically perceptual unit that can be inferred by visual image features, without any prior knowledge of what the camera wearer is actually doing.}
In \cref{fig:duration} a histogram with the duration (in minutes) of the resulting segments is shown. We can observe a wide variability in duration. Most of the segments (around 65\%) have relatively short durations of 15 minutes or less, but there also exist several long events (around 5\%) with a duration longer than 100 minutes.

The dataset contains a total of 48,717 images, divided in 1,339 events (or image sequences) and 3,991 captions, and has an average of 3 captions per event. It was divided in training, validation and test splits making sure that all the sequences from the same day should belong to the same data split. The division results in the figures depicted in \cref{table:dataset}.

\begin{table}[!h]
\centering
\caption{\label{table:dataset}Figures of the EDUB-SegDesc dataset, according to each partition, training, validation and test.}

\begin{tabular}{p{3cm}p{2cm}p{2cm}p{2cm}p{2cm}}
\toprule
EDUB-SegDesc  & Training & Validation & Test & Total \\ \midrule
\#days & 39 & 7 & 9 & 55 \\
\#images & 32,664 & 7,301 & 8,752 & 48,717  \\
\#segments & 889 & 204 & 246 & 1,339 \\
\#descriptions & 2,652 & 598 & 741 & 3,991 \\ 
\bottomrule
\end{tabular}
\end{table}

\begin{wrapfigure}{r}{0.4\textwidth}
	\centering
	\includegraphics[width=0.4\textwidth]{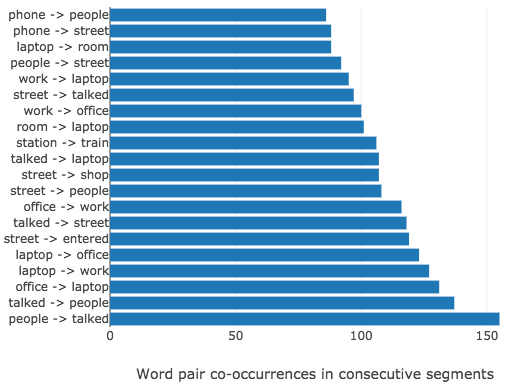}
	\caption{\label{fig:word_pairs} Number of co-occurrences in consecutive segments of several manually chosen keywords. Word pairs are shown in the following format: \texttt{word1 -\textgreater word2}, where \texttt{word1} appears in the past event and \texttt{word2} appears in the current event.}
\end{wrapfigure}

In ~\cref{fig:word_pairs} we show the number of co-occurrences in consecutive segments from some manually chosen keywords. This highlights the natural relationships found in consecutive events. Some notable examples of concepts appearing in consecutive events include: 'people' in past events followed by 'talked' in current events (social events); 'laptop' followed by 'work' (work-related events); 'street' followed by 'entered' (going from an outdoors to an indoors environment); 'station' followed by 'train' (transport-related events) or 'phone' followed by 'street' (events related to using the mobile phone on the street).

\begin{figure}[htp]

\centering
\includegraphics[width=.3\textwidth]{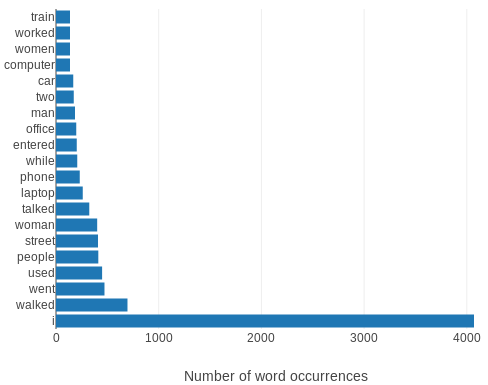}\hfill
\includegraphics[width=.34\textwidth]{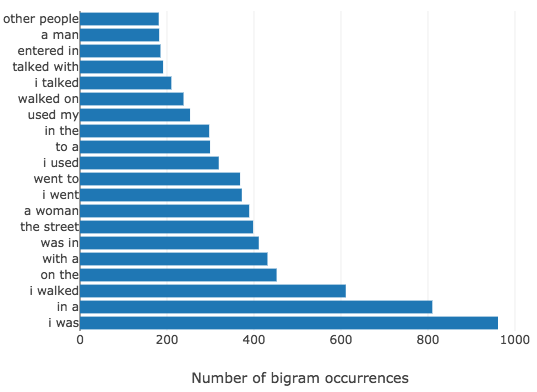}\hfill
\includegraphics[width=.34\textwidth]{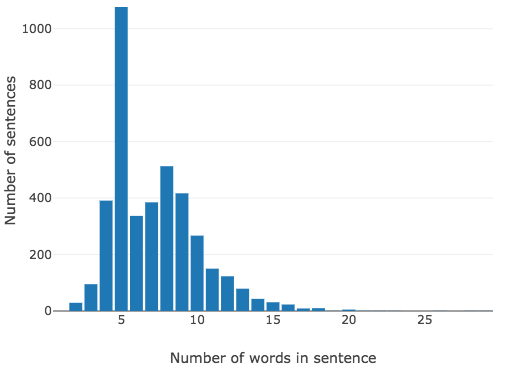}

\caption{Word-related statistics of the dataset. Occurrences of the most common words (left), occurrences of the most common bigrams (center), and histogram of number of words in all the sentences (right).}
\label{fig:word_statistics}
\end{figure}

\cref{fig:word_statistics} shows several word-related statistics: Occurrences of the most common words and bigrams and histogram sentence lengths. Note that the number of appearances of the word 'I' is very high both in the single word and the bigram counts given the egocentric nature of the dataset. Other commonly occurring words and bigrams are the ones related to events where the user is 'walking' and/or on the 'street'. Curiously, there is a certain bias in the number of words contained in the annotations, where a considerable number of the sentences are composed of 5 words. Some examples of 5-word sentences are 'I went to my office', 'I worked with my laptop', or 'I walked on the street'.

\begin{figure}[!ht]
	\centering
	\includegraphics[width=0.7\textwidth]{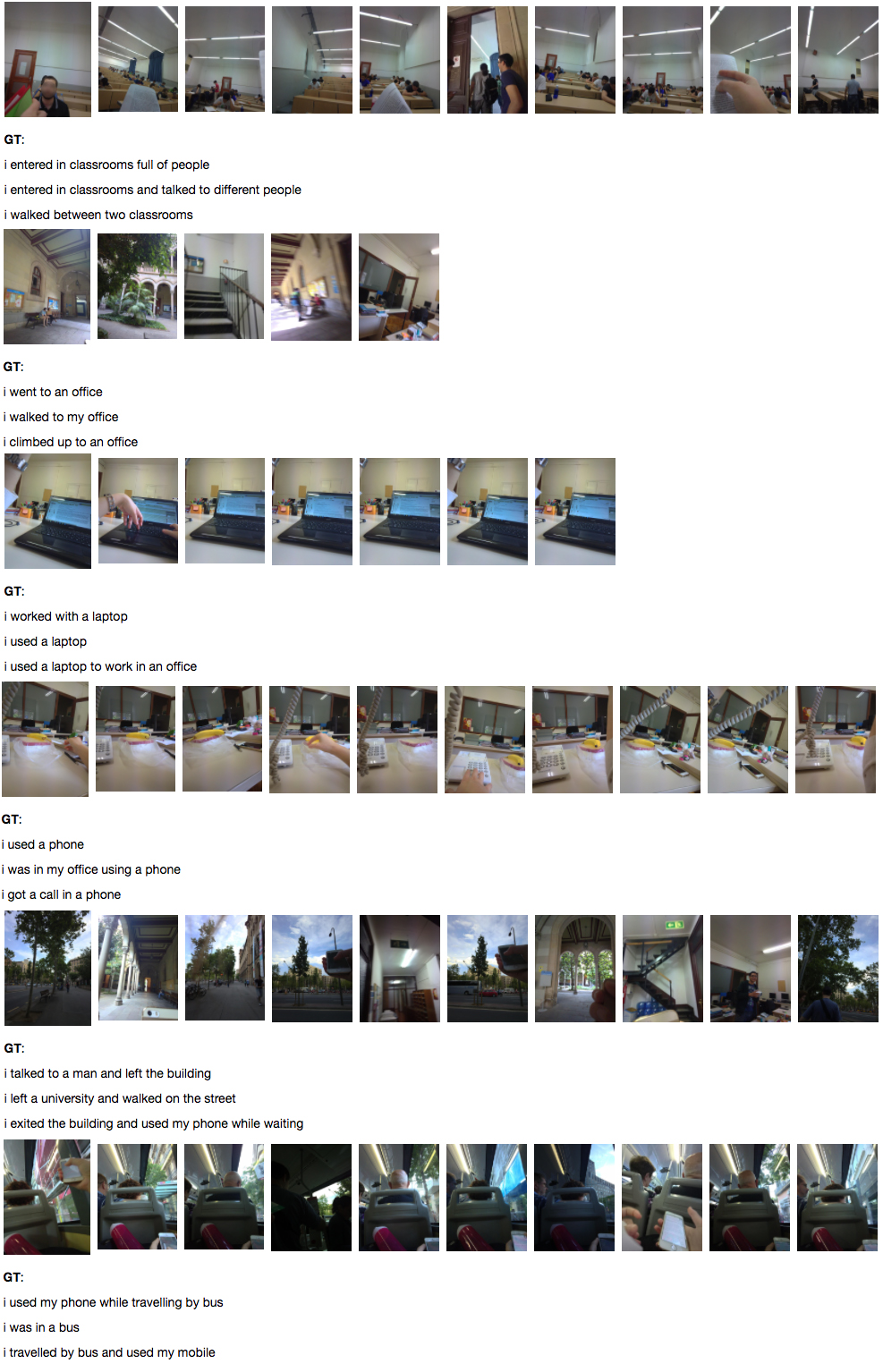}
	\caption{\label{fig:dataset_sample} Subset of a day from the dataset EDUB-SegDesc. We show some consecutive events and their respective GT sentences. The difficulty of the problem is highlighted by the dynamism and instability of the scene, when the user is moving. Particularly difficult examples can be seen in the second event: the GT sentences explain that the user is heading to his/her office, but this aspect is only manifested in the last image of the sequence.}
	\vspace*{0.2cm}
\end{figure}

Finally, we show some examples of the events and sentences contained in the dataset (\cref{fig:dataset_sample}). The low temporal resolution of the camera used (2 frames per minute) becomes clear in dynamic events, where the user moves and this causes to have highly variable environments. This fact, together with the limited information present in some of the images highlights the difficulty of the problem. 

The EDUB-SegDesc dataset was specifically acquired and labeled for the purpose of developing a model for describing and understanding all the events appearing along the day of a person. Thus, the main application of using egocentric sequences for textual descriptions generation is providing a memory aid for MCI patients.

\section{Experiments and results} 
\label{sec:results}

In this section we set up the experimental environment. We also define the metrics employed to evaluate it. Finally, we show the results obtained.

\subsection{Metrics and evaluation}

We evaluated our proposal using standard image and video captioning metrics. We used the COCO-Caption evaluation package \citep{Chen15} and computed three metrics:

\textbf{BLEU}~\citep{Papineni02}:
compares the ratio of n-gram structures that are shared between the system hypotheses and the reference sentences. 
We report BLEU-4 in our results.
%


\textbf{METEOR}~\citep{Lavie09}: this metric was introduced to solve the lack of the recall component when computing BLEU. It
computes the F1 score of precision and recall between hypotheses and references. In addition, it considers exact, stemmed, synonyms and paraphrase matches. 
%
%

\textbf{CIDEr}~\citep{Vedantam15}:
similarly to BLEU, it computes the number of matching n-grams, but penalizing n-grams frequently found in the training set. The CIDEr metric ranges from  0 (minimum quality) to 10 (maximum quality).

%
%

\subsection{Experimental setup}

All the neural models that we compare were built with the Keras\footnote{www.keras.io} and Theano~\citep{Theano16} libraries. We release the source code of our implementation for future comparisons\footnote{https://github.com/MarcBS/TMA}. The full model was jointly trained end-to-end on the EDUB-SegDesc dataset, except for the CNN, which was already pre-trained for object detection on ImageNet~\citep{Russakovsky14} and remained static during the training of our model. 

The main hyper-parameters of the model were selected according to the analysis reported in the video description task~\citep{Peris16}. Moreover, since we conducted experiments with pre-trained models, we must keep fixed some hyper-parameters. Therefore, we used word embeddings of size 301, the encoder BLSTM had 717 units in the forward and 717 more units in the backward layers and the decoder LSTM had 484 units. The initial state of the decoder LSTM was initialized with the hyperbolic tangent of the mean of the video features obtained by the encoder~\citep{Xu15}. We took at most 26 frames evenly distributed from each complete sequence of our dataset.

We trained our model by stochastic gradient descent (SGD). After doing experiments either using Adam or Adadelta optimizers, we observed that for some model variations, the optimal performance was reached either using Adadelta~\citep{Zeiler12} with a learning rate of 1.0 and without learning rate decay; or using Adam~\citep{kingma2014adam} with an initial learning rate of 0.001 and a decay of 0.995 at the end of every epoch. Thus, we report the best results on each case. During training, the norm of the gradients was clipped to 10~\citep{Pascanu13}.

In order to prevent over-fitting, we used batch normalization~\citep{Ioffe15}. Contrarily to other works, we observed that the use of dropout~\citep{Srivastava14} combined with batch normalization, produced better performance,
($p=0.5$). We also applied weight decay ($10^{-4}$) and Gaussian noise ($\sigma=10^{-2}$) to the non-recurrent weights. 

We set our batch size to 64 and used an early stop criterion on the validation set based on BLEU-4, setting our patience to 20 and checking the performance each 50 updates. The size of the beam during the search was 10.

In order to minimize the influence of randomness in our results, mostly due to the weights random initialization, each experiment was run 5 times, and reported the median value of such runs. The model was trained either from scratch, exclusively using the data from the EDUB-SegDesc dataset or reusing the pre-trained weights from certain layers that were learned in different model variations. We studied the inclusion of word embeddings, obtained using the skip-gram model from \citet{Mikolov13} and trained on part of Google News dataset. We also tested training the decoder as a language model on the 1 Billion words dataset~\citep{Chelba13}, but results in both cases were not better. Finally, we also tested reusing the pre-trained weights from the video captioning model from~\citet{Peris16} (trained on the MSVD dataset published by \citet{Chen11}), which proved to improve the results under certain model configurations.
The Microsoft Research Video Description Corpus (MSVD) dataset~\citep{Chen11} contains 1,970 short clips from YouTube annotated by different users, accounting for more than 80,000 training samples. In terms of coverage, approximately the 98\% of the words from EDUB-SegDesc are present in the MSVD dataset.

\subsection{Results}

In this section, we study the influence on the final performance of different architectural proposals. First, we show the performance of a classical video caption system and study the influence of several pre-training methods. Next, we study some of the architectural choices that led to the best TMA system. Finally, we compare the state of the art with our best proposals proposal based on the influence that information from previous events has on the system.

\cref{table:results-video} shows the results obtained when using different pre-training techniques for the language model. In all the experiments we tackle the problem as classical video captioning without incorporating information from previous events. The captioning model is the same as in~\citet{Peris16}.

As shown in ~\cref{table:results-video}, the inclusion of word2vec vectors worsens the performance of the system in terms of BLEU and METEOR. We hypothesize that this is due to the different domains on which the word embeddings are trained. The word2vec vectors were trained on a more general domain and, therefore, their capabilities cannot be exploited to the full in our problem. 

If we use the parameters learned from MSVD data, the BLEU score is also lowered, although in a lower extent than with word2vec. Nevertheless, the performance of the MSVD model in terms of METEOR and CIDEr is increased. 

\begin{table}[h]
	\caption{\label{table:results-video}EDUB-SegDesc validation and test set results for different pre-training techniques on the language model. ABiViRNet refers to the video captioning system from~\citet{Peris16}. BLEU and METEOR metrics are given in percentage. We compare the basic model trained from scratch with the same model with certain pre-trained components. \#params denotes the number of parameters to estimate, given in millions.}
	\centering
	\footnotesize
	\renewcommand{\arraystretch}{1.1}
	\begin{tabular}{{lllllllr}}
		\toprule
        & \multicolumn{3}{c}{Validation} & \multicolumn{3}{c}{Test} & \#params \\  
        \cmidrule(lr){2-4} \cmidrule(lr){5-7}
		& BLEU-4 & METEOR & CIDEr & BLEU-4 & METEOR & CIDEr & \\ 
		\midrule
		ABiViRNet & 31.2 & 21.3 & 0.97 & 29.6 & 20.3 & 0.79 & 27.3M\\
		ABiViRNet + word2vec & 30.1 & 21.0 & 1.04 & 26.0 & 20.1 & 0.90 & 27.3M  \\
		ABiViRNet + MSVD & 32.5 & 22.0 & 1.11 & 28.5 & 21.2 & 0.89 & 35.1M\\
		\bottomrule
	\end{tabular}
\end{table}

In \cref{table:results-extra} we report a further set of comparisons that show several additional tests that did not provide good results. In the first part of the table we compare several models either using or not dropout. It appears that, combining the use of dropout, batch normalization and Gaussian noise in the same model clearly helps obtaining better results in all the cases. The sole use of dropout as regularization strategy produced even worse results.  
In the second part of the table we compare either using all the images available in the dataset or removing all non-informative images \citep{lidon2015semantic} (i.e. images which are either dark, blurry or that point to the sky or ground without showing any object). 

\begin{table}[h]
	\caption{\label{table:results-extra}EDUB-SegDesc validation and test set results for several additional configurations tested that did not provide good enough results. For each bad result we report its counter example configuration with better results. BLEU and METEOR metrics are given in percentage. \#params denotes the number of parameters to estimate, given in millions. The best results for each measure are shown in boldface. Results with the symbol * were obtained with the Adam optimizer instead of Adadelta.}
	\centering
	\footnotesize
	\renewcommand{\arraystretch}{1.1}
	\begin{tabular}{lllllllr}
		\toprule
        & \multicolumn{3}{c}{Validation} & \multicolumn{3}{c}{Test} & \#params \\ \cmidrule(lr){2-4} \cmidrule(lr){5-7}
		& BLEU-4 & METEOR & CIDEr & BLEU-4 & METEOR & CIDEr &  \\ 
        \midrule
        ABiViRNet no dropout & 29.3 & 21.6 &  1.09 & 24.8 & 18.6 & 0.80 & 27.3M\\
        ABiViRNet & 31.2 & 21.3 & 0.97 & 29.6 & 20.3 & 0.79 & 27.3M \\
        ABiViRNet + MSVD no dropout & 30.2 & 21.0 & 1.05 & 27.9 & 19.4 & 0.89 & 35.1M\\
        ABiViRNet + MSVD & 32.5 & 22.0 & 1.11 & 28.5 & 21.2 & 0.89 & 35.1M\\
        TMA previous-caption no dropout & 31.9 & 21.6 & 1.02 & 28.1 & 20.1 & 0.94 & 39.1M \\
		TMA previous-caption & 32.6 & 21.7 & 1.04 & 30.6 & 20.4 & 0.90 & 39.1M \\
		\midrule
        TMA previous-video + MSVD - NonInfo* & 31.7 & 21.7 & 1.13 & 30.3 & 21.5 & 0.99 & 51.0M \\
		TMA previous-video + MSVD* & 35.4 & 23.5 & 1.18 & 31.9 & 22.1 & 1.07 & 51.0M \\
		\bottomrule
	\end{tabular}
\end{table}


\begin{table}[h]
	\caption{\label{table:results-previous-info}EDUB-SegDesc validation and test set results for variations of our TMA model compared to the state-of-the-art, which do not consider information from previous events. We use either the previous caption, the previous sequence of images (video) or both for making the current prediction. BLEU and METEOR metrics are given in percentage. \#params denotes the number of parameters to estimate, given in millions. The best results for each measure are shown in boldface. Results with the symbol * were obtained with the Adam optimizer instead of Adadelta.}
	\centering
	\footnotesize
	\renewcommand{\arraystretch}{1.1}
	\begin{tabular}{lllllllr}
		\toprule
        & \multicolumn{3}{c}{Validation} & \multicolumn{3}{c}{Test} & \#params \\ \cmidrule(lr){2-4} \cmidrule(lr){5-7}
		& BLEU-4 & METEOR & CIDEr & BLEU-4 & METEOR & CIDEr &  \\ 
        \midrule
        Enc-Dec Global~\citep{Yao15} & 30.1 & 21.9 & 1.02 & 28.1 & 20.8 & 0.88 & 44.4M\\
        hLSTMat \citep{song2017hierarchical} & 31.6 & 23.3 & 1.28 & 25.6 & 20.8& 0.88 & ---\\
        ABiViRNet \citep{Peris16} & 31.2 & 21.3 & 0.97 & 29.6 & 20.3 & 0.79 & 27.3M \\
        DeepSeek \citep{goeldeepseek} &  33.6 & 24.1 & 1.26  & 27.9 & 21.4 & 0.99 &  27.2M \\
		\midrule
		TMA previous-caption & 32.6 & 21.7 & 1.04 & 30.6 & 20.4 & 0.90 & 39.1M \\
		TMA previous-caption + MSVD & 33.9 & 23.5 & 1.21 & 28.3 & 21.3 & 0.94 & 46.9M \\
		\midrule
		TMA previous-video* & 34.4 & 23.7 &  1.21 & 31.0 & 21.4 & 1.01 & 43.3M \\
		TMA previous-video + MSVD* & 35.4 & 23.5 & 1.18 & \textbf{31.9} & \textbf{22.1} & \textbf{1.07} & 51.0M \\
		\midrule
		TMA previous-video-caption & \textbf{36.4} & \textbf{24.6} & \textbf{1.29} & 30.4  & 21.8 & 1.00 & 55.1M \\
		TMA previous-video-caption + MSVD* & 34.0 & 23.0 & 1.19 & 29.7 & \textbf{22.1} & 0.93 & 62.8M \\
		\bottomrule
	\end{tabular}
\end{table}

The effect of including information from previous events is shown in \cref{table:results-previous-info}. For comparison, we also include the only similar approach in the egocentric video captioning literature, \textit{DeepSeek}~\citep{goeldeepseek}. This model consists of a non-attentional BLSTM encoder with two layers. The encoder feeds a similar decoder to the one from~\citet{Yao15}. We performed additional tests with several state-of-the-art models for video captioning: \textit{Enc-Dec Global}~\citep{Yao15}, \textit{hLSTMat}~\citep{song2017hierarchical} and \textit{ABiViRNet}~\citep{Peris16}. Note that none of them considers long-term temporal information from different events.

As detailed in~\cref{sec:mutliLSTM}, we tested our TMA model introducing the previous caption, the previous sequence of images or both previous caption and images. As before, we distinguish between training from scratch with the EDUB-SegDesc dataset or start from the weights learned with MSVD pre-trained model. Given the results observed in~\cref{table:results-video}, we drop the use of word2vec word embeddings.

According to~\cref{table:results-previous-info}, if we compare the results obtained by the state of the art methods (lines 1-4), which only consider the current sequence of images, to the different configurations of our method (lines 5-10), we can see that, in most of the cases, our method outperforms the rest. This behaviour can be explained by the inclusion of information from previous events in the TMA model. Since it considers a broader context, it is able to better understand the given event, and therefore, generally succeeds at increasing the performance of the system. Furthermore, considering the characteristics of egocentric lifelogging photostreams, the data provided by a single event often lacks enough information to easily understand what is happening. Thus, providing context from previous events usually improves the captioning results. On the other hand, some state of the art methods outperform certain configurations of our model (see line 3 BLEU-4, and line 4 METEOR and CIDEr on the test set). We understand that this phenomenon can occur, because although providing context from previous events can usually be useful, in some cases noise could be introduced for two reasons: 1) the error of the predictions of previous events can be propagated, and 2) some consecutive events could lack any semantical relationship or have a very low number of samples in the training set. 

 \begin{wrapfigure}{R}{0.5\textwidth}
	\centering
	\includegraphics[width=0.5\textwidth]{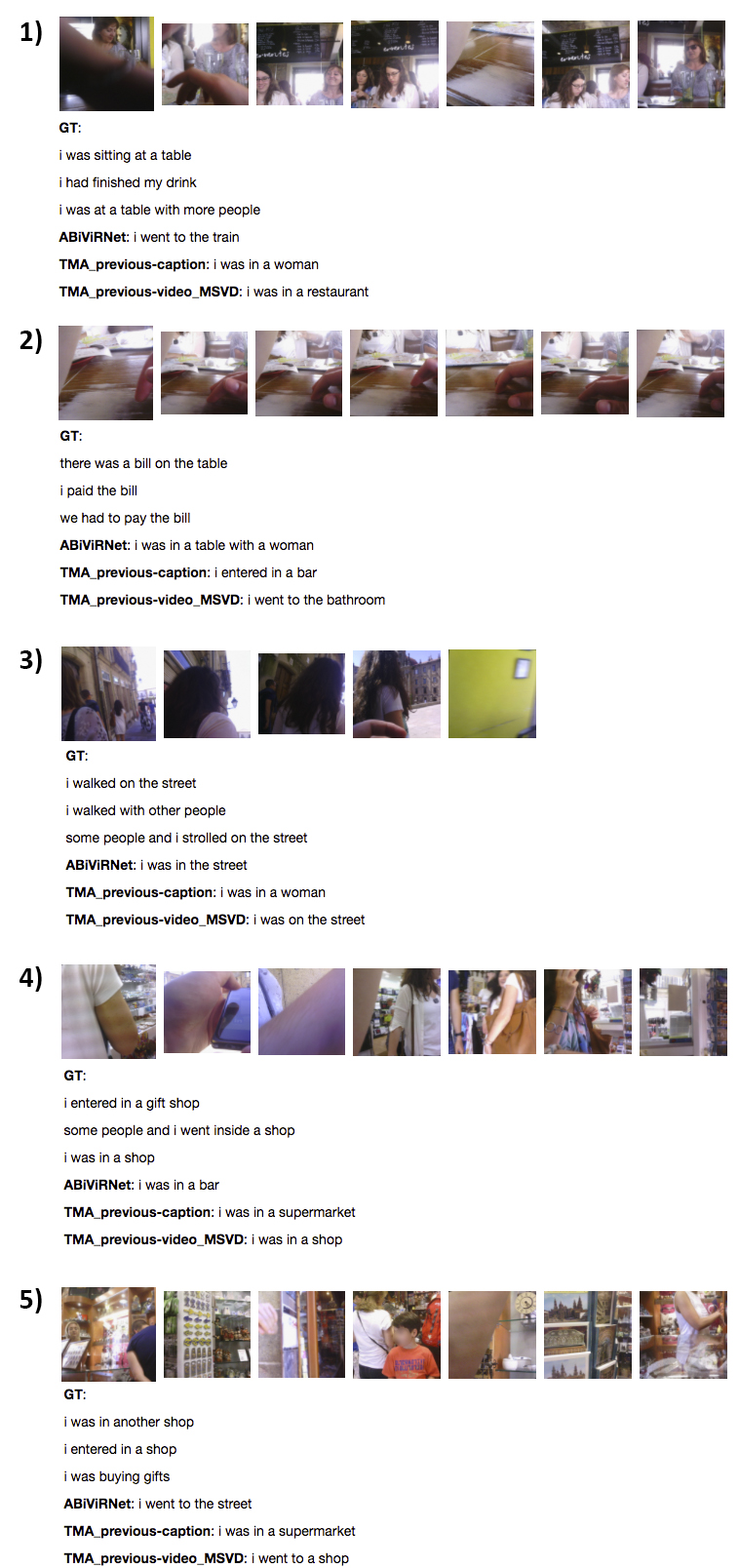}
	\caption{\label{fig:results_1} Comparison of the results obtained by the baseline method ABiViRNet and two of our proposals TMA previous-caption and TMA previous-video + MSVD. A set of consecutive events from the test split is shown.}
\end{wrapfigure}

Considering the differences between different configurations of our TMA model, the inclusion of the previous video event as input yields the best generalization results during test evaluation. The previous caption also enhances the system, but in a lower extent. On the other hand, the inclusion of both previous frames and caption produced better results, but only in the validation set. This phenomenon could imply that although the model has a greater potential, it is also more complex considering the number of parameters, which produces a quicker over-fitting on the training data and the consequent impact produced by the model selection of the validation set  \citep{reunanen2003overfitting}.

The results obtained with the TMA model show that taking into account the previous video event is more effective than considering the previous caption. This effect is partially produced because, at test time, we use as input an output previously generated by the model. Obviously, this output may be erroneous. Therefore, it may lead to the introduction of noise to the system and to error propagation. 

From~\cref{table:results-video} and \cref{table:results-previous-info}, it can be concluded that, when fine-tuning from the MSVD model, the METEOR scores are always better than when starting from scratch, but BLEU-4 scores are lowered. This is probably due to vocabulary differences from both tasks. The model pre-trained tends to produce captions with the structure that it learned from the MSVD dataset, while the model trained from scratch on the EDUB-SegDesc generates ad-hoc captions for the task at hand. Since BLEU-4 is based on n-gram counts, the latter model generally obtains larger values than the first one, because its captions are more literal. Since the METEOR metric stems and employs synonyms, it is less rigid than BLEU-4. Therefore, pre-trained models are able to score better than those trained from scratch.

\section{Discussion}
\label{sec:discussion}

In this section, we review some illustrative examples, in order to understand the weaknesses and strengths of our model, as well as the major difficulties appeared in the task at hand.

\cref{fig:results_1} shows some examples of the predictions produced by our model on consecutive events in the test set. We can observe the influence that the previous event had in the captioning of the second sample. In sample \#2, the \textit{TMA previous-video} model is aware that the user was previously in a restaurant. This conditions the model, which generates the caption \textit{"I went to the bathroom"}, which is a likely action when a person is in a restaurant. The \textit{ABiViRNet} model is unable to capture this information.

The influence of previous actions can also be observed in events \#5 and \#6. At event \#5, the user is shopping and in the next event, he/she continues shopping. The \textit{TMA previous-video} model can distinguish the shopping action occurred in event \#6 due to incorporation of the visual information from the previous one. On the other hand, the \textit{ABiViRNet} model is not able to do this inference and deduces that the user is in the street. Similarly, the \textit{TMA previous-caption} model is more prone to error propagation: when fails to understand a certain event, it is more likely to be also wrong in the following one. For example, in sample \#4 the model generates \textit{``I was in a supermarket"}, which leads the model to fail also on the consecutive event \#5.

\begin{wrapfigure}{R}{0.5\textwidth}
	\centering
	\includegraphics[width=0.5\textwidth]{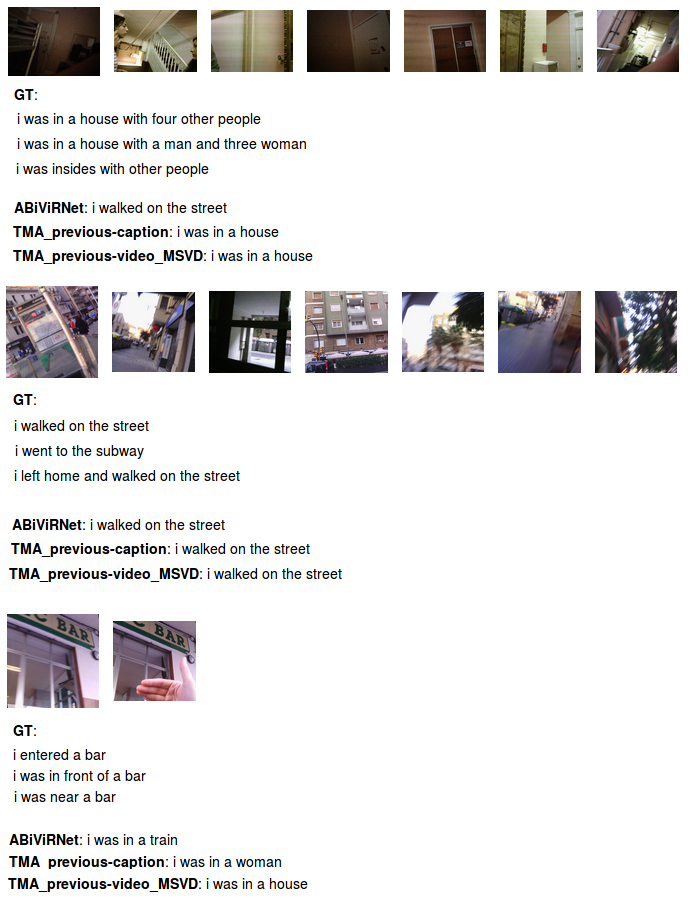}
	\caption{\label{fig:first_event_samples} Comparison of the results obtained by the baseline method ABiViRNet and two of our proposals TMA previous-caption and TMA previous-video + MSVD. Events corresponding to the start of the day are shown.}
\end{wrapfigure}

\cref{fig:first_event_samples} depicts some examples of initial events of different days. We can see that, in the case of TMA models, the fact that information from previous events can not be used in this particular cases does not influence the model and obtains better or comparable results than non-TMA models.

\begin{figure}[!ht]
\centering
\begin{minipage}{.5\textwidth}
  \centering
  \includegraphics[width=\textwidth]{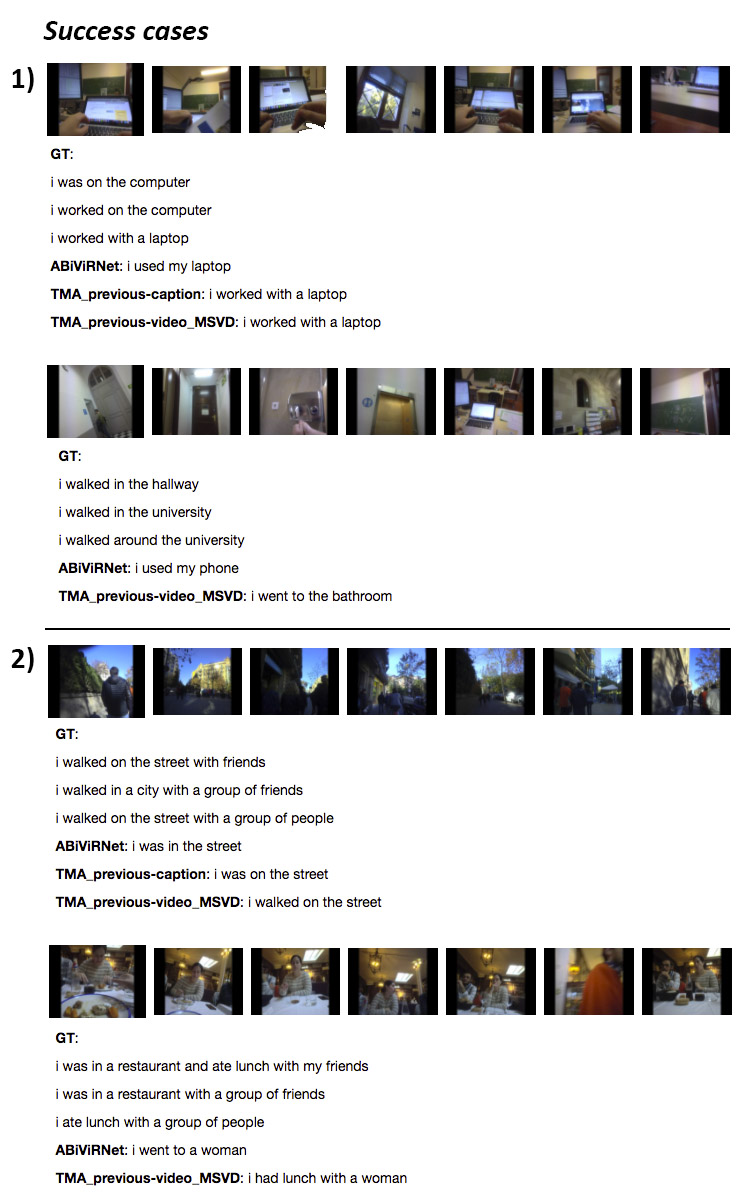}
\end{minipage}%
\begin{minipage}{.5\textwidth}
  \centering
  \includegraphics[width=0.875\textwidth]{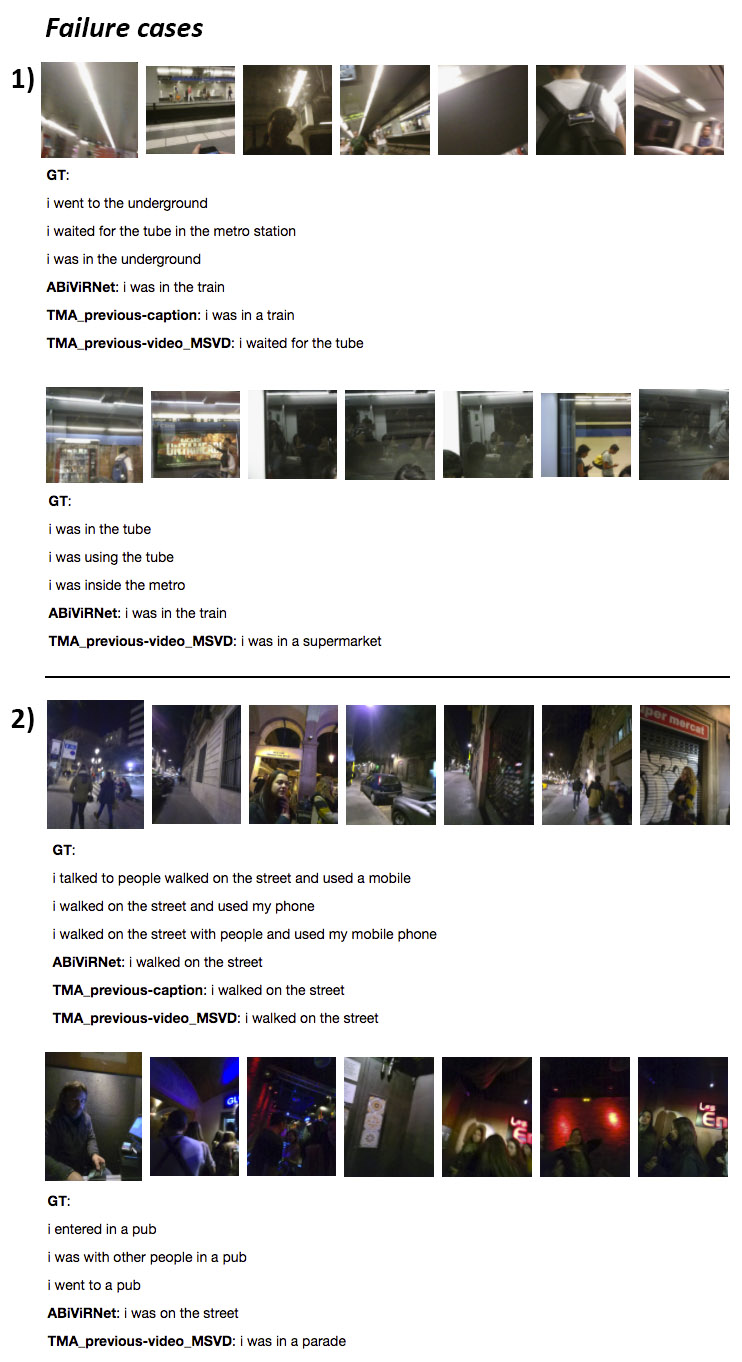}
\end{minipage}
\caption{\label{fig:results_2} Two success cases (left) and two failure cases (right) of our model \textit{TMA previous-video + MSVD}. All the examples belong to the test set. For each case, the top sequence is the previous event and the bottom sequence is the  event that we are predicting. In success case \#1, we can see that our method correctly recognizes that the user went to the bathroom even though it is not specified in the GT sentences and it is not straightforward to see in the images (note the third image using a hand drier). The two samples at the right exemplify cases in which our model fails due to certain images appearing in the sequences. For instance, in the failure case \#1 our model considers that the user is in a supermarket due to the first image, where a vending machine appears. In failure case \#2, our model considers that the user is seeing a parade due to the multiple people and colored lights appearing in the images.}
\end{figure}

\cref{fig:results_2} shows additional examples of successes/failures of the system. In the success case \#1, we can see that the model is able to understand that the user went to the bathroom, although it is not specifically described in the GT and it is hard to distinguish just looking at the images (note the third image where the user is drying his hands). In addition, this exemplifies one of the major challenges occurred within the event captioning problem: during an event, many situations may occur. The system may focus only on part of the event, resulting in captions that, although correct, are not found in the GT.

In the success case \#2, we can see that both \textit{ABiViRNet} and \textit{TMA previous-video} model correctly describe the first event. But as we move to the next one, the first model is unable to detect that the user is having lunch, while the TMA model is capable to infer that the event has changed. Therefore, it incorporates this information and correctly guesses that the user is having lunch.

The right examples show failure cases caused under certain conditions in the input images. In the first failure case, the model infers that the user is in a supermarket due to the vending machine in the first image. In the second failure case, the model interprets the images as a parade due to the multiple persons and colored lights appearing.

\section{Conclusions and future work} \label{sec:conclusions}

 In this work, we addressed the challenging problem of egocentric captioning as a video description task considering a natural characteristic of this problem: since the egocentric sequences were captured consecutively along a day, there exists a relationship between a given situation and the previous one. We aimed to include such dependency in an automatic captioning system. For doing this, we developed a natural extension of LSTM networks, able to deal with multiple inputs, even when coming from different modalities.

 For assessing our proposal, we also constructed a dataset for egocentric captioning. Image sequences were obtained using a wearable camera and were manually segmented and annotated.  Both source code and dataset are made publicly available. We carried out an automatic evaluation of the system, with clear results: the inclusion of previous information effectively enhances the performance of the system.
  
If we have available the sequence of events of a full day, we could introduce not only previous events to the system, but also the following ones. Since the TMA model defined in this work supports an arbitrary number of inputs, the inclusion of the context coming from following events becomes natural. Therefore, the captioning results could be refined, not only by the previous, but also by incorporating the following information. Furthermore, we could aim to look further than the immediately previous event, incorporating longer-term memory. Very recently, \citet{Kaiser17} proposed a memory-augmented network, able to learn very long-term relationships. We plan to  test whether such architectures can deal with our problem in an effective way.
 
Additionally, in a hypothetical real application, the inclusion of an interactive correction process could be considered. Interactive neural systems provide encouraging results~\citep{Knowles16,Peris16b} in the machine translation field. In such systems, the user corrects the errors committed by the system, the system takes into account these corrections and changes its outputs, to produce a hopefully better caption. 
Finally, closely related with interactivity, another interesting extension of this work is the application of online learning techniques to the egocentric captioning pipeline. This would allow the development of better user-tailored and adaptive captioning systems, definitely more useful to the final user.

\section*{Acknowledgments}
This work was partially founded by TIN2015-66951-C2, SGR 1219, CERCA,  Grant 20141510 (Marat\'{o} TV3), PrometeoII/2014/030 and R-MIPRCV network (TIN2014-54728-REDC). Petia Radeva is partially founded by ICREA Academia'2014. Marc Bola\~nos is partially founded by an FPU fellowship. We gratefully acknowledge the support of NVIDIA Corporation with the donation of a Titan X GPU used for this research. The funders had no role in the study design, data collection, analysis, and preparation of the manuscript. 

\newpage

\bibliographystyle{elsarticle-harv}
\bibliography{main}

\end{document}